\documentclass[10pt,twocolumn,letterpaper]{article}

\usepackage{titling}
\usepackage{iccv}
\usepackage{times}
\usepackage{epsfig}
\usepackage{graphicx}
\usepackage{amsmath}
\usepackage{amssymb}

\usepackage{bbold}
\usepackage{wasysym}
\usepackage{soul}
\usepackage{bm}
\usepackage{stmaryrd}

\usepackage{multirow}
\usepackage{tabularx}
\usepackage{booktabs}
\usepackage{subcaption}
\usepackage{graphbox}
\usepackage{textcomp}

\DeclareMathOperator{\Diag}{Diag}
\DeclareMathOperator{\softmax}{softmax}
\newcommand{\tens}[1]{
\bm{\mathcal{#1}}
}
\newcommand{\mat}[1]{\bm{#1}}
\newcommand{\p}{\mathbf{p}}
\newcommand{\q}{\mathbf{q}}
\newcommand{\vv}{\mathbf{v}}

\newcommand{\z}{\mathbf{z}}
\newcommand{\y}{\mathbf{y}}
\newcommand{\tq}{\mathbf{\tilde{q}}}
\newcommand{\tv}{\mathbf{\tilde{v}}}
\newcolumntype{C}{>{\centering\arraybackslash}X}
\usepackage[breaklinks=true,bookmarks=false]{hyperref}

\iccvfinalcopy % *** Uncomment this line for the final submission

\def\iccvPaperID{****} % *** Enter the ICCV Paper ID here

\ificcvfinal\fi
\begin{document}
%%%%%%%%% TITLE
\title{MUTAN: Multimodal Tucker Fusion for Visual Question Answering}

\author{Hedi Ben-younes $^{1,2}$ \thanks{Equal contribution}
% For a paper whose authors are all at the same institution,
% omit the following lines up until the closing ``}''.
% Additional authors and addresses can be added with ``\and'',
% just like the second author.
% To save space, use either the email address or home page, not both
\qquad 
R\'emi Cadene $^1$\footnotemark[1] 
% {\tt\small secondauthor@i2.org}
\qquad  
Matthieu Cord $^1$  
% {\tt\small secondauthor@i2.org}
\qquad 
Nicolas Thome $^3$ 
\\ $^1$ Sorbonne Universit\'es, UPMC Univ Paris 06, CNRS, LIP6 UMR 7606, 4 place Jussieu, 75005 Paris
\\ $^2$ Heuritech, 248 rue du Faubourg Saint-Antoine, 75012 Paris 
\\ $^3$ Conservatoire National des Arts et M\'etiers
\\  {\tt\small hedi.ben-younes@lip6.fr, remi.cadene@lip6.fr, matthieu.cord@lip6.fr, nicolas.thome@cnam.fr}
}
\maketitle

\vspace{-0.2cm}
\begin{abstract}
Bilinear models provide an appealing framework for mixing and merging information in Visual Question Answering (VQA) tasks.
They help to learn high level associations between question meaning and visual concepts in the image, but they suffer from huge dimensionality issues.  

We introduce MUTAN, a multimodal tensor-based Tucker decomposition to efficiently parametrize bilinear interactions between visual and textual representations. 
Additionally to the Tucker framework, we design a low-rank matrix-based decomposition to explicitly constrain the interaction rank. With MUTAN, we control the complexity of the merging 
scheme while keeping nice interpretable fusion relations. We show how our MUTAN model generalizes some of the latest VQA architectures, providing state-of-the-art results.
\end{abstract}
\vspace{-0.2cm}
%%%%%%%%% BODY TEXT
\section{Introduction}

Multimodal representation learning for text and image has been extensively studied in recent years. Currently,  the most popular task is certainly Visual Question Answering (VQA) \cite{Malinowski2014, VQA}. VQA is a complex multimodal task which aims at answering a question about an image. 
A specific benchmark has been first proposed \cite{Malinowski2014}, and large scale datasets have been recently collected~\cite{DBLP:conf/nips/RenKZ15,VQA,zhu2016cvpr}, enabeling the development of more powerful models. 

To solve this problem, precise image and text models are required and, most importantly, high level interactions between these two modalities have to be carefully encoded into the model in order to provide the correct answer.  
This projection from the unimodal spaces to a multimodal one is supposed to extract and model the relevant correlations between the two spaces. Besides, the model must have the ability to understand the full scene, focus its attention on the relevant visual regions and discard the useless information regarding the question.

\begin{figure}[htbp] 
\includegraphics[align=c,width=\linewidth]{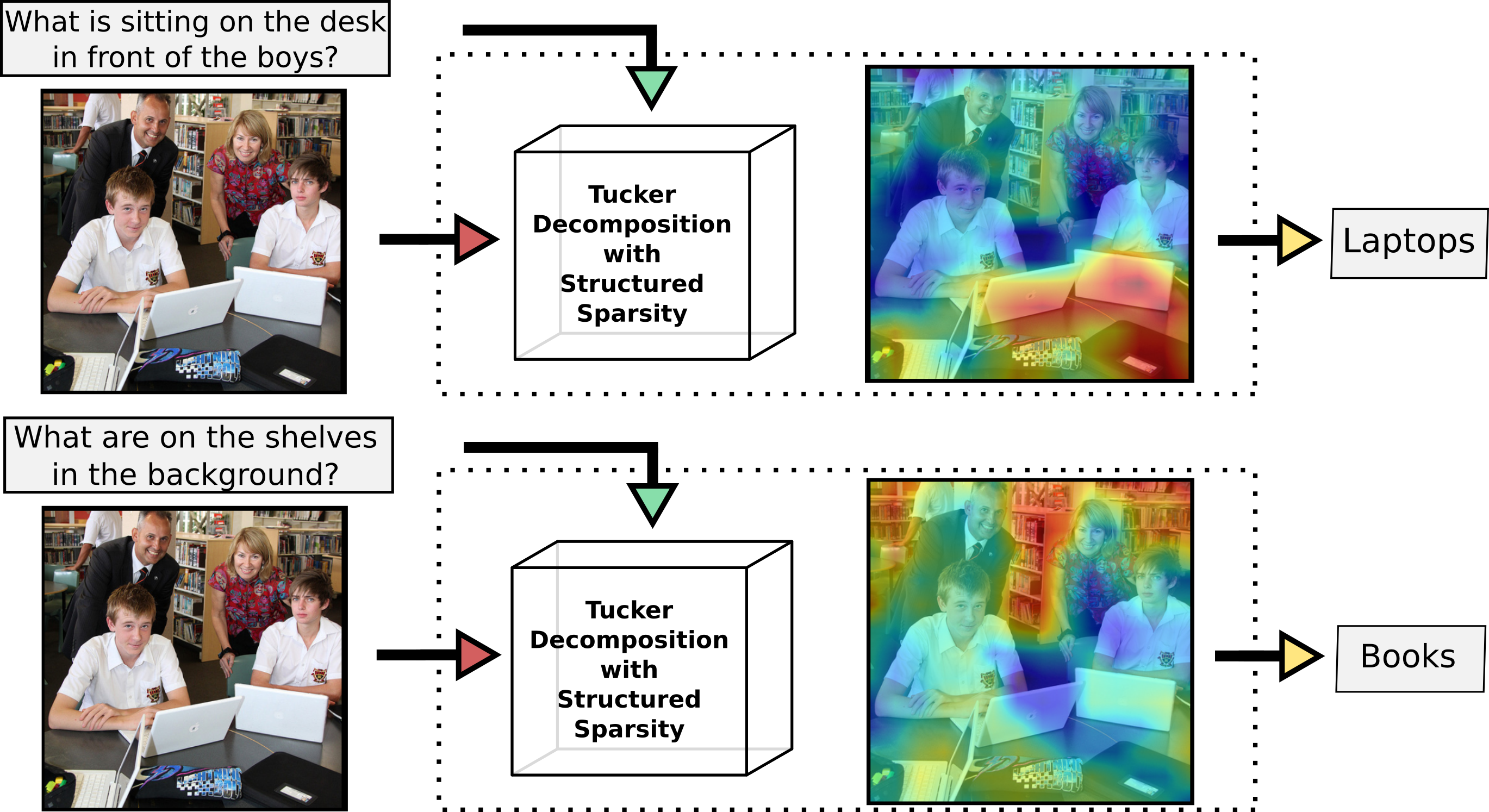}
\label{visu_intro}
\caption{
The proposed MUTAN model uses a Tucker decomposition of the image/question correlation tensor, which enables modeling rich and accurate multi-modal interactions.
~For the same input image, we show the result of the MUTAN fusion process when integrated into an attention mechanism \cite{icml2015_xuc15}: we can see that the regions with larger attention scores (in red) indicate a very fine understanding of the image and question contents,
enabling MUTAN to properly answer the question (see detailed maps in experiments section).}
\end{figure}

Bilinear models are powerful approaches for the fusion problem in VQA because they encode full second-order interactions. They currently hold state-of the-art performances~\cite{fukui16mcb, Kim2017}. 
The main issue with these bilinear models is related to the number of parameters, which quickly becomes intractable with respect to the input and output dimensions.
Therefore, current bilinear approaches must be simplified or approximated by reducing the model complexity: in~\cite{fukui16mcb}, the authors sacrifice trainability by using a handcrafted multi-modal projection, 
while a global  tensor rank constraint is applied in~\cite{Kim2017}, reducing correlations to a simple element-wise product.  

\begin{figure*}[htbp]
\begin{center}
\includegraphics[width=\linewidth]{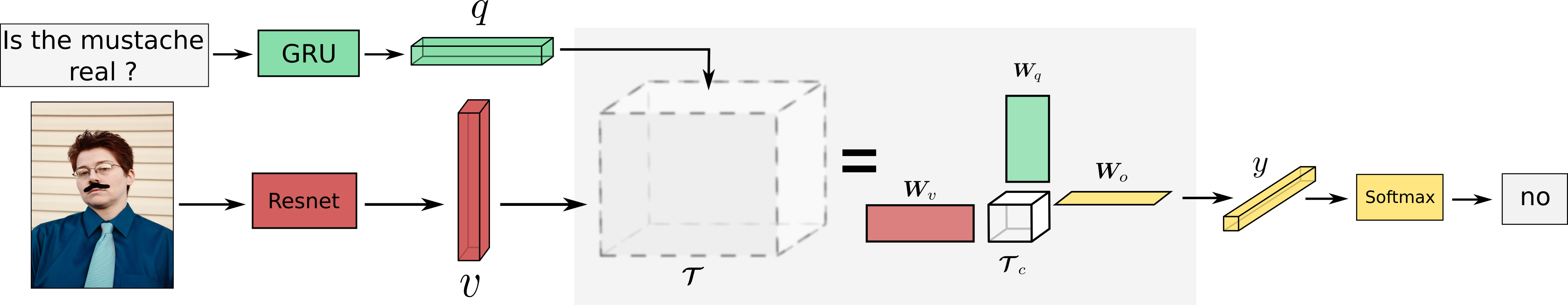}
\caption{\label{whole_model} MUTAN fusion scheme for global Visual QA. 
The prediction is modeled as a bilinear interaction between visual and linguistic features, parametrized by the tensor $\tens{T}$. 
In MUTAN, we factorise the tensor $\tens{T}$ using a Tucker decomposition, resulting in an architecture with three intra-modal matrices $\mat{W}_q$, $\mat{W}_v$ and $\mat{W}_o$, 
and a smaller tensor $\tens{T}_c$. The complexity of $\tens{T}_c$ is controlled \textit{via} a structured sparsity constraint on the slice matrices of the tensor.} 
\end{center}
\end{figure*}

In this work, we introduce a new architecture called MUTAN (Figure~\ref{whole_model}), which focuses on modeling fine and rich interactions between image and textual modalities.  
Our approach is based on a Tucker decomposition~ \cite{Tucker1966} of the correlation tensor, which is able to represent full bilinear interactions, while maintaining the size of the model tractable. 
The resulting scheme allows us to explicitly control the model complexity, and to choose an accurate and interpretable repartition of the learnable parameters. 

In the next section, we provide more details on related VQA works and highlight our contributions. The MUTAN fusion model, based on a Tucker decomposition, is presented in section~\ref{sec:model}, and 
successful experiments  are reported in section~\ref{sec:expes}.

%-------------------------------------------------------------------------
\section{Related work}
The main task in multimodal visual and textual analysis aims at learning an alignment between feature spaces \cite{Yan_2015_CVPR,TACL325,MaLSL15}. 
Thus, the recent task of image captioning aims at generating linguistic descriptions of images
\cite{VinyalsTBE16,icml2014c2_kiros14,icml2015_xuc15}. Instead of explicitly learning an alignment between two spaces, 
the goal of VQA \cite{VQA,Malinowski2014} is to merge both modalities in order to decide which answer is correct.
This problem requires modeling very precise correlations between the image and the question representations.

\noindent {\bf Attention.} Attention mechanisms \cite{icml2015_xuc15} have been a real breakthrough in multimodal systems, and are fundamental for VQA models to obtain the best possible results. \cite{YangHGDS16} propose to stack multiple question-guided attention mechanisms, each one looking at different regions of the image. 

 \cite{shih2016wtl} and \cite{NIPS2016_6261} extract bounding boxes in the image and score each one of them according to the textual features. 
In \cite{LuYBP16}, word features  are aggregated with an attention mechanism guided by the image regions and, equivalently, the region visual features are aggregated into one global image embedding. This co-attentional framework uses concatenations and sum pooling to merge all the components. On the contrary, \cite{fukui16mcb} and \cite{Kim2017} developed their own fusion methods that they use for global and attention-based strategies. 

In this paper, we use the attentional modeling, proposed in  \cite{fukui16mcb}, as a tool that we integrate in our new fusion strategy for both the global fusion and the attentional modeling.

\noindent {\bf Fusion strategies.}
Early works have modeled interactions between multiple modalities with first order interactions. The IMG+BOW model in \cite{DBLP:conf/nips/RenKZ15} is the first 
to use a concatenation to merge a global image representation with a question embedding, obtained by summing all the learnt word embeddings from the question. 
In \cite{shih2016wtl}, (image, question, answer) triplets are scored in an attentional framework. Each local feature is given a score corresponding to its similarity with textual features. 
These scores are used to weight region multimodal embeddings, obtained from a concatenation between the region's visual features and the textual embeddings. 
The hierarchical co-attention network \cite{LuYBP16}, after extracting multiple textual and visual features, merges them with concatenations and sums.

Second order models are a more powerful way to model interactions between two embedding spaces. Bilinear interactions have shown great success in deep learning for fine-grained classification \cite{lin2015bilinear}, and Multimodal language modeling  \cite{icml2014c2_kiros14}.
In VQA, a simple element-wise product between the two vectors is performed in \cite{VQA}. \cite{kim2016b} also uses  an element-wise product
in a more complex iterative global merging scheme.  In \cite{NIPS2016_6261}, they use the element-wise product aggregation in an attentional framework. To go deeper in bilinear interactions,  Multimodal Compact Bilinear pooling (MCB)~\cite{fukui16mcb} uses an outer product $\q \otimes \vv$ between visual $\vv$ and textual $\q$ embeddings. The count-sketch projection \cite{Charikar:2002:FFI:646255.684566} $\Psi$
is used to project $\q \otimes \vv$ on a lower dimensional space. Interestingly, nice count-sketch properties are capitalized to compute the projection
without having to explicitly compute the outer product. 
However, interaction parameters in MCB are fixed
by the count-sketch projection (randomly chosen in $\{0; -1; 1\}$), limiting its expressive power for modeling complex interactions between image and questions. In contrast, our approach is able to model 
rich second order interaction with learned parameters. 

In the recent Multimodal Low-rank Bilinear (MLB) pooling work~\cite{Kim2017}, full bilinear interactions 
between image and question spaces are parametrized by a tensor. Again, to limit the number of free parameters, this tensor is constrained to be of low rank $r$. The MLB strategy reaches state-of-the-art performances on the  well-known VQA database~\cite{VQA}. Despite these impressive results, the low rank tensor structure is equivalent to a projection of both visual and question representations into a common $r$-dimensional space,  and to compute simple element-wise product interactions in this space. 
MLB is thus essentially designed to learn a powerful mono-modal embedding for text and image modalities, but relies on a simple 
fusion scheme in this space. 

In this work, we introduce MUTAN, a multimodal fusion scheme based on bilinear interactions between modalities. 
To control the number of model parameters, MUTAN reduces the size of the mono-modal embeddings, while modeling their interaction as accurately as possible with a full bilinear fusion scheme. 
Our submission therefore encompasses the following contributions:

-- New fusion scheme for VQA relying on a Tucker tensor-based decomposition, consisting in a factorization into three matrices and a core tensor. 
We show that the MUTAN fusion scheme generalizes the latest bilinear models, \textit{i.e.} MCB~\cite{fukui16mcb} and MLB~\cite{Kim2017}, while having more expressive power. 
 
 -- Additional structured sparsity constraint the core tensor to further control the number of model parameters. This acts as a regularizer during training and prevents over-fitting, giving us more flexibility to adjust the input/output projections.
 
 -- State-of-the-art results on the most widely used dataset for Visual QA~\cite{VQA}. 
We also show that MUTAN outperforms MCB~\cite{fukui16mcb} and MLB~\cite{Kim2017} in the same setting, and that performances can be further improved when combined with MLB, validating the complementarity potential
between the two approaches.

%------------------------------------------------------------------------
\section{MUTAN Model}
\label{sec:model}
Our method deals with the problem of Visual Question Answering (VQA). 
In VQA, one is given a  question $q \in \mathcal{Q}$ about an image $v \in \mathcal{I}$, and the goal is to provide a meaningful answer.
During training, we aim at learning a model such that
the predicted answer $\hat{\mathbf{a}}$ matches the correct one $\mathbf{a}^\star$. 
More formally, denoting as $\Theta$ the whole set of parameters of the model, the predicted output $\hat{\mathbf{a}}$ can be written as:

\begin{equation}
\label{eq:vqaprediction}
\hat{a} = \arg \underset{a \in \mathcal{A}} \max ~ p_{\Theta}\left( a | v,q \right)  
\end{equation}

The general architecture of the proposed approach is shown in Figure~\ref{whole_model}. 
As commonly done in VQA, images $v$ and questions $q$ are firstly embedded into vectors and the output is represented as a classification vector $\y$ . In this work, we use a fully convolutional neural network~\cite{He2015} (ResNet-152) to describe the image content,
and a GRU recurrent network \cite{Kiros2015, ChoMBB14} for the question, yielding representations $\vv \in \mathbb{R}^{d_v}$ for the image and $\q \in \mathbb{R}^{d_q}$ for the question.
Vision and language representations $\vv$ and $\q$ are then fused using the operator $\tens{T}$ (explained below) to produce a vector $\y$, providing (through a softmax function) the final answer in Eq.~(\ref{eq:vqaprediction}). This global merging scheme is also embedded into a visual attention-based mechanism \cite{Kim2017} to provide our final MUTAN architecture.

\paragraph{Fusion and Bilinear models}
The issue of merging visual and linguistic information is crucial in VQA. Complex and 
high-level interactions between textual meaning in the question and visual concepts in the image have to be extracted to provide a meaningful answer. 

Bilinear models~\cite{fukui16mcb,Kim2017} are recent powerful solutions to the fusion problem, since they encode fully-parametrized bilinear interactions between the vectors $\q$ and $\vv$: 

\begin{equation}
 \label{bilinear} 
\y = \left( \tens{T} \times_1  \q \right) \times_2  \vv
\end{equation}
with the full tensor $\tens{T} \in \mathbb{R}^{d_q \times d_v \times |\mathcal{A}|}$% and $\y \in \mathbb{R}^{|\mattcal{A}|}$
, and the operator $\times_i$ designing the \emph{i-mode} product between a tensor and a matrix (here a vector). 

Despite their appealing modeling power, fully-parametrized bilinear interactions quickly become intractable in VQA, because the size of the full tensor is  
prohibitive using common dimensions for textual, visual and output spaces. 
For example, with $d_v \approx d_q \approx 2048$ and $|\mathcal{A}| \approx 2000$, the number of free parameters in the tensor $\tens{T}$ is $\sim 10^{10}$.
Such a huge number of free parameters is a problem both for learning and for GPU memory consumption\footnote{A tensor with 8 billion float32 scalars approximately needs 32Go to be stored, while top-grade GPUs hold about 24Go each.}. 

 In MUTAN, we factorize the full tensor $\tens{T}$ using a Tucker decomposition. 
We also propose to complete our decomposition by structuring the second tensor $\tens{T}_c$ (see gray box in Fig.~\ref{whole_model}) in order to keep flexibility over the input/output dimensions while keeping the number of parameters tractable.

\label{bilinear_pb}
\subsection{Tucker decomposition}
The Tucker decomposition \cite{Tucker1966} of a 3-way tensor $\tens{T}~\in~\mathbb{R}^{d_q \times d_v \times |\mathcal{A}|}$ expresses $\tens{T}$ as a tensor product between \emph{factor matrices} $\mat{W}_q, \mat{W}_v$ and $\mat{W}_o$, and a \emph{core tensor} $\tens{T}_{c}$ in such a way that:
\begin{equation}
\tens{T} = \left( \left( \tens{T}_{c}\times_1 \mat{W}_q \right) \times_2 \mat{W}_v \right) \times_3 \mat{W}_o \label{tucker}
\end{equation}
with $\mat{W}_q \in \mathbb{R}^{d_q \times t_q}$, $\mat{W}_v \in \mathbb{R}^{d_v \times t_v}$ and $\mat{W}_o \in \mathbb{R}^{|\mathcal{A}| \times t_o}$, and $\tens{T}_c \in \mathbb{R}^{t_q \times t_v \times t_o}$. Interestingly, Eq.~(\ref{tucker}) states that the weights in $\tens{T}$ are functions of a restricted number of parameters $\forall i\in [1,d_q], j \in [1,d_v], k \in [1,d_o]$:
\vspace*{-0.3cm}
\begin{equation*}
\tens{T}[i,j,k] = \hspace{-1cm}\sum_{\substack{l \in [1,t_q], m \in [1,t_v], n \in [1,t_o]}} \hspace{-1cm}\tens{T}_c[l,m,n] \mat{W}_q[i,l] \mat{W}_v[j,m] \mat{W}_o[k,n]
\end{equation*}
$\tens{T}$ is usually summarized as $\tens{T} = \llbracket \tens{T}_c; \mat{W}_q,\mat{W}_v,\mat{W}_o \rrbracket$.
A comprehensive discussion on Tucker decomposition and tensor analysis may be found in \cite{Kolda:2009:TDA:1655228.1655230}.

\subsection{Multimodal Tucker Fusion} \label{mtf}
As we parametrize the weights of the tensor $\tens{T}$ with its Tucker decomposition of the Eq.~(\ref{tucker}), we can rewrite Eq.~(\ref{bilinear}) as follows:
\begin{align}
y &= \left( \left(\tens{T}_{c} \times_1 \left(\q^\top \mat{W}_q \right)\right) \times_2 \left(\vv^\top \mat{W}_v \right)\right) \times_3 \mat{W}_{o}
\end{align}
This is strictly equivalent to encode a full bilinear interaction of projections of $q$ and $v$ into a latent pair representation $\z$, and to use this latent code to predict the correct answer. If we define $\tq = \q^ \top \mat{W}_q \in \mathbb{R}^{t_q}$ and $\tv = \vv^ \top \mat{W}_v \in \mathbb{R}^{t_v}$, we have:
\begin{align} \label{core_bilinear}
\z = (\tens{T}_c \times_1 \tq) \times_2 \tv\in \mathbb{R}^{t_o} 
\end{align}
$\z$ is projected into the prediction space $\y = \z^\top\mat{W}_o \in \mathbb{R}^{|\mathcal{A}|}$ and $\p = \softmax(\y)$. In our experiments, we use non-linearities $\tq = \tanh(\q^\top \mat{W}_q)$ and $\tv = \tanh(\vv^\top \mat{W}_v)$ in the fusion, as in \cite{Kim2017}, providing slightly better results. The multimodal Tucker fusion is depicted in Figure~\ref{whole_model}.

\paragraph{Interpretation}
Using the Tucker decomposition, we have separated $\tens{T}$ into four components, each having a specific role in the modeling. Matrices $\mat{W}_q$ and $\mat{W}_v$ project the question and the image vectors into spaces of respective dimensions $t_q$ and $t_v$. These dimensions directly impact the modeling complexity that will be allowed for each modality. The higher $t_q$ (resp. $t_v$) will be, the more complex the question (resp. image) modeling will be. Tensor $\tens{T}_c$ is used to model interactions between $\tq$ and $\tv$. It learns a projection from all the correlations $\tq[i] \tv[j]$ to a vector $\z$ of size $t_o$. This dimension controls the complexity allowed for the \emph{interactions} between modalities. Finally, the matrix $\mat{W}_o$ scores this pair embedding $\z$ 
for each class in $\mathcal{A}$.

\subsection{Tensor sparsity} \label{section:ssparsity}
To further balance between expressivity and complexity of the interactions modeling, we introduce a structured sparsity constraint based on the rank of the slice matrices in $\tens{T}_c$.
When we perform the $t_o$ bilinear combinations between $\tq$ and $\tv$ of Eq.~(\ref{core_bilinear}), each dimension $k \in \llbracket 1,t_o \rrbracket$ in $\z$ can be written as:
\begin{equation}
\z[k] = \tq^\top \tens{T}_c[:,:,k] \tv
 \label{ssparsity}
\end{equation}
The correlations between elements of $\tq$ and $\tv$ are weighted by the  parameters of $\tens{T}_c[:,:,k]$. We might benefit from the introduction of a structure in each of these slices. This structure can be expressed in terms of rank constraints on the slices of $\tens{T}_c$. We impose the rank of each slice to be equal to a constant R. Thus we express each slice $\tens{T}_c[:,:,k]$ as a sum of R rank one matrices:
\begin{equation}
\tens{T}_c[:,:,k] = \sum_{r=1}^R \mathbf{m}_r^k \otimes \mathbf{n}_r^{k\top}
\end{equation}
with $\mathbf{m}_r^k \in \mathbb{R}^{t_q}$ and $\mathbf{n}_r^k \in \mathbb{R}^{t_v}$
Eq.~(\ref{ssparsity}) becomes:
\begin{equation}
z[k] = \sum_{r=1}^R \left( \tq^\top \mathbf{m}_r^k \right) \left( \tv^\top \mathbf{n}_r^k \right)
\end{equation}
We can define R matrices $\mat{M}_r~\in~\mathbb{R}^{t_q~\times~t_o}$ (resp. $\mat{N}_r~\in~\mathbb{R}^{t_v~\times~t_o}$) such as $\forall k \in \llbracket 1, d_o \rrbracket, M_r[:,k] = \mathbf{m}_r^k$ (resp. $N_r[:,k] = \mathbf{n}_r^k$). The structured sparsity on $\tens{T}_c$ can then be written as:
\begin{gather}
\z = \sum_{r=1}^R \z_r \\
\z_r = (\tq^\top \mat{M}_r) \ast (\tv^\top \mat{N}_r )
\end{gather}

\paragraph{Interpretation}
Adding this rank constraint on $\tens{T}_c$ leads to expressing the output vector $\z$ as a sum over $R$ vectors $\z_r$. To obtain each of these vectors, we project $\tq$ and $\tv$ into a common space and merge them with an elementwise product. Thus, we can interpret $\z$ as modeling an OR interaction over multiple AND gates ($R$ in MUTAN) between projections of $\tq$ and $\tv$. $\z[k]$ can described in terms of logical operators as: 
\begin{gather}
\z_r[k] = \left( \tq\text{ similar to }\mathbf{m}_r^k \right)\text{ AND }\left( \tv\text{ similar to }\mathbf{n}_r^k\right) \\
\z[k] = \z_1[k]\text{ OR }...\text{ OR }\z_R[k]
\end{gather}
This decomposition gives a very clear insight of how the fusion is carried out in our MUTAN model. In our experiments, we will show how different $r$'s in $\llbracket 1,  R\rrbracket$ behave, depending on the type of question. We will exhibit some cases where some $r$'s specialize over specific question types.
\subsection{Model Unification and Discussion}\label{uniformization}
\begin{figure*}[t]
\centering
  \begin{subfigure}{0.3\linewidth}
	\centering
    \includegraphics[height=3cm]{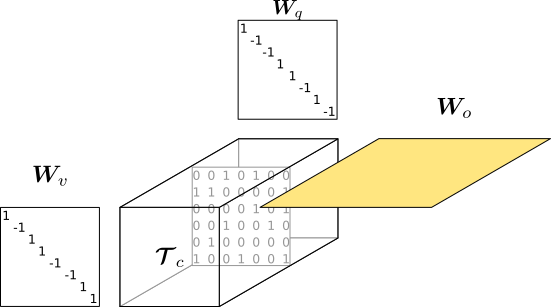}
	\caption{\label{bilinear_compare:a} MCB}
  \end{subfigure}
  \hspace{1px}
  \begin{subfigure}{0.3\linewidth}
  \centering
    \includegraphics[height=3cm]{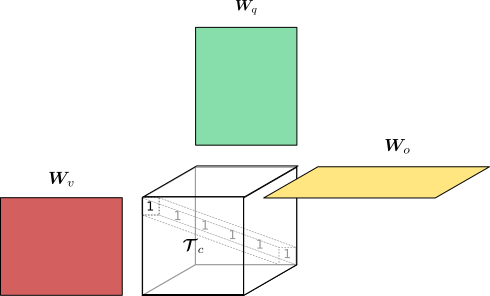} 
    \caption{\label{bilinear_compare:b} MLB} 
  \end{subfigure} 
  \begin{subfigure}{0.3\linewidth}
  \centering
    \includegraphics[height=3cm]{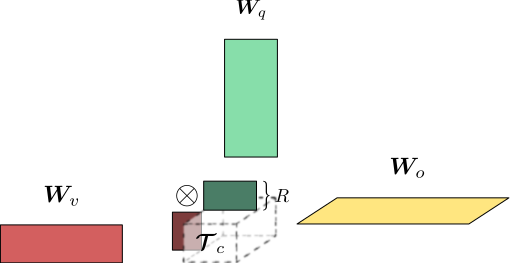}  
    \caption{\label{bilinear_compare:c} MUTAN}
  \end{subfigure}
  \caption{\label{bilinear_compare} Tensor design strategies. (a) MCB:  $\mat{W}_q$ and $\mat{W}_v$ are fixed diagonal matrices, $\tens{T}^c$ is a sparse fixed tensor, only the output factor matrix $\mat{W}_o$ is learnt; (b) MLB: the 3 factor matrices are learnt but the core tensor is $\tens{T}^c$  set to identity; 
 (c)~MUTAN: $\mat{W}_q$, $\mat{W}_v$, $\mat{W}_o$ and $\tens{T}^c$ are learnt. The  full bilinear interaction $\tens{T}^c$ is structured with a low-rank (R) decomposition.
}
  \label{fig7} 
\end{figure*}

In this subsection, we show how two state of the art models, namely Multimodal Low-rank bilinear pooling \cite{Kim2017} (MLB) and Multimodal Compact Bilinear pooling \cite{fukui16mcb} (MCB), can be seen as special cases of our Multimodal Tucker Fusion. Each of these models use a different type of bilinear interaction between $\q$ and $\vv$, hence instantiating a specific parametrization of the weight tensor $\tens{T}$. These parameterizations actually consist  in a Tucker decomposition with specific constraints on the elements $\tens{T}_c,\mat{W}_q,\mat{W}_v$ and $\mat{W}_o$. More importantly, when we cast MCB and MLB into the framework of Tucker decompositions, we show that the structural constraints imposed by these two models state that some parameters are fixed, while they are free to be learnt in our full Tucker fusion. This is illustrated in Figure~\ref{bilinear_compare}. We show in color the learnt parameters.

\subsubsection{Multimodal Compact Bilinear (MCB)}
We can show that the Multimodal Compact Bilinear pooling \cite{fukui16mcb} can be written as a bilinear model where the weight tensor $\tens{T}^{mcb}$ is decomposed into its Tucker decomposition, with specific structures on the decompositions' elements. The intramodal projection matrices $\mat{W}_q^{mcb}$ and $\mat{W}_v^{mcb}$ are diagonal matrices where the non-zero coefficients take their values in $\{-1;1\}$:
$\mat{W}_q^{mcb} = \mat{\Diag{}} (\mathbf{s}_q)$ and $\mat{W}_v^{mcb} = \mat{\Diag{}} (\mathbf{s}_v)$, where $\mathbf{s}_q \in \mathbb{R}^{d_q}$ and $\mathbf{s}_v \in \mathbb{R}^{d_v}$ are random vectors sampled at the instanciation of the model but kept fixed afterwards. The core tensor $\tens{T}_c$ is sparse and its values follow the rule: $\tens{T}_c^{mcb}[i,j,k] = 1$  if  $h(i,j) = k$ (and $0$  else), where $h: \llbracket 1,d_q \rrbracket \times \llbracket 1,d_v \rrbracket \rightarrow \llbracket 1, d_o\rrbracket$ is randomly sampled at the beginning of training and no longer changed.

As was noticed in \cite{Kim2017}, all the learnt parameters in MCB are located \emph{after} the fusion. The combinations of dimensions from $\q$ and from $\vv$ that are supposed to interact with each other are randomly sampled beforehand (through $h$). To compensate for the fact of fixing the parameters $\mathbf{s}_q, \mathbf{s}_v$ and $h$, they must set a very high $t_o$ dimension (typically 16,000). This set of combinations is taken as a feature vector for classification. 

\subsubsection{Multimodal Low-rank Bilinear (MLB)}

The low-rank bilinear interaction corresponds to a canonical decomposition of the tensor $\tens{T}$ such as its rank is equal to R. It is well-known that the low-rank decomposition of a tensor is a special case of the Tucker decomposition, such as $\tens{T}^{mlb} = \llbracket \tens{I}_R; \mat{W}_q, \mat{W}_v, \mat{W}_o \rrbracket$
where $t_q = t_v = t_o = R$. Two major constraints are imposed when reducing Tucker decomposition to low-rank decomposition. First, the three dimensions $t_q, t_v$ and $t_o$ are structurally set to be equal. The dimension of the space in which a modality is projected ($t_q$ and $t_v$) quantifies the model's complexity. Our intuition is that since the image and language spaces are different, they may require to be modeled with different levels of complexity, hence different projection dimensions. The second constraint is on the core tensor, which is set to be the identity. A dimension $k$ of $\tq^{mlb}$ is only allowed to interact with the same dimension of $\tv^{mlb}$, which might be restrictive. We will experimentally show the beneficial effect of removing these constraints.

We would like to point out the differences between MLB and the structured sparsity per slice presented in \ref{section:ssparsity}. There are two main differences between the two approaches. First, our rank reduction is made on the core tensor of the Tucker decomposition $\tens{T}_c$, while in MLB they constrain the rank of the global tensor $\tens{T}$. This lets us keep different dimensionalities for the projected vectors $\tq$ and $\tv$. The second difference is we do not reduce the tensor on the third mode, but only on the first two modes corresponding to the image and question modalities. The implicit parameters in $\tens{T}_c$ are correlated \emph{inside} a mode-3 slice but independent \emph{between} the slices.

\section{Experiments}
\label{sec:expes}
\paragraph{VQA Dataset}
The VQA dataset \cite{VQA} is built over images of MSCOCO \cite{mscoco}, where each image was manually annotated with 3 questions. Each one of these questions is then answered by 10 annotators, yielding a list of 10 ground-truth answers.
The dataset is composed of 248,349 pairs (image, question) for the training set, 121,512 for validation and 244,302 for testing. The ground truth answers are given for the \textit{train} and \textit{val} splits, and one must submit their predictions to an evaluation server to get the scores on \textit{test-std} split. Note that the evaluation server makes it possible to submit multiple models per day on \textit{test-dev}, which is a subsample of \textit{test-std}. The whole submission on \textit{test-std} can only be done \emph{five} times. We focus on the open-ended task, where the ground truth answers are given in free natural language phrases. This dataset comes with its evaluation metric, presented in \cite{VQA}. When the model predicts an answer for a visual question, the VQA accuracy is given by: 
\begin{equation}
\min \left( 1, \frac{\text{\# humans that provided that answer}}{3} \right)
\label{acc_vqa}
\end{equation}
If the predicted answer appears at least 3 times in the ground truth answers, the accuracy for this example is considered to be 1. Intuitively, this metrics takes into account the consensus between annotators.

\paragraph{MUTAN Setup}
We first resize our images to be of size $(448,448)$. We use ResNet152 \cite{He2015} as our visual feature extractor, 
which produces feature maps of size $14\times14\times2048$. 
We keep the $14\times14$ tiling when attention models are used (section ~\ref{final_system}). Otherwise, the image is represented as the average of $14 \times 14$ vectors at the output of the CNN (section~\ref{sec:compafusion}).
To represent questions, we use a GRU \cite{ChoMBB14} initialized with the parameters of a pretrained Skip-thoughts model \cite{Kiros2015}. 
Each model is trained to predict the most common answer in the 10 annotated responses.
$|\mathcal{A}|$ is fixed to the $2000$ most frequent answers as in \cite{Kim2017}, and we train our model using ADAM \cite{KingmaB14} (see details in supplementary material).

\subsection{Fusion Scheme Comparison}
\label{sec:compafusion}

To point out the performance variation due to the fusion modules, we first compare MUTAN to state-of-the-art bilinear models, under the same experimental framework. 
We do not use attention models here. 
Several merging scheme results are presented in Table~\ref{fusion_compare:testdev}: Concat denotes a baseline where $\vv$ and $\q$ are merged by simply concatenating them. For MCB \cite{fukui16mcb} and MLB \cite{Kim2017}, we use the available code \footnote{\url{https://github.com/jnhwkim/cbp}} \footnote{\url{https://github.com/jnhwkim/MulLowBiVQA}} to train models on the same visual and linguistic features. We choose an output dimension of 16,000 for MCB and 1,200 for MLB, as indicated in the respective articles. MUTAN\_noR designates the MUTAN model without the rank sparsity constraint. We choose all the projection dimensions to be equal to each other: $t_q = t_v = t_o = 160$. These parameters are chosen considering the results on \textit{val} split. Finally, our MUTAN \footnote{\url{https://github.com/cadene/vqa.pytorch}} designates the full Tucker decomposition with rank sparsity strategy. We choose all the projection dimensions to be equal to each other: $t_q = t_v = t_o = 360$, and a rank $R = 10$. These parameters were chosen so that MUTAN and MUTAN\_noR have the same number of parameters.
 As we can see in Table~\ref{fusion_compare:testdev}, MUTAN\_noR performs better than MLB, which validates the fact that modeling full bilinear interactions between low dimensional projections yields a more powerful representation than having strong mono-modal transformations with a simple fusion scheme (element-wise product). With the structured sparsity, MUTAN obtains the best results, validating our intuition of having a nice tradeoff between the projection dimensions and a reasonable number of useful bilinear interaction parameters in the core tensor $\tens{T}_c$.
Finally, a naive late fusion MUTAN+MLB further improves performances (about +1pt on {\em test-dev}). It validates the complementarity between the two types of tensor decomposition.

\begin{table}[]
\begin{tabularx}{\columnwidth}{lX*{4}{X}X}
&&\multicolumn{4}{c}{\textit{test-dev}} & \textit{val} \\
\cmidrule{3-6} \cmidrule{7-7}
Model & $\Theta$ & Y/N & No. & Other & All & All\\ 
\hline
Concat & 8.9 & 79.25 & 36.18 & 46.69 & 58.91 & 56.92\\ 
MCB & 32 & 80.81 & 35.91 & 46.43 & 59.40 & 57.39 \\ 
 MLB & 7.7 & \textbf{82.02} & 36.61 & 46.65 & 60.08 & 57.91\\ 
 MUTAN\_noR & 4.9 & 81.44 & 36.42 & 46.86 & 59.92 & 57.94\\ 
MUTAN & 4.9 & 81.45 & \textbf{37.32} & \textbf{47.17} & \textbf{60.17} & \textbf{58.16}\\
\hline
\hline
\textcolor{black}{MUTAN+MLB} & 17.5 & 82.29 & 37.27 & 48.23 & 61.02 & 58.76 \\

\bottomrule
\end{tabularx}
\caption{\label{fusion_compare:testdev} Comparison between different fusion under the same setup on the \textit{test-dev} split. $\Theta$ indicates the number of learnable parameters (in million).}  
\end{table}

\subsection{State-of-the-art comparison} 
\label{final_system}

To compare the performance of the proposed approach to state-of-the-art works, we associate the MUTAN fusion with recently introduced techniques for VQA, which are described below.

\vspace{-0.4cm}
\paragraph{Attention mechanism}
We use the same kind of multi-glimpse attention mechanisms as the ones presented in \cite{fukui16mcb} and \cite{Kim2017}. We use MUTAN to score the region embeddings according to the question vector, and compute a global visual vector as a sum pooling weighted by these scores.
\vspace{-0.4cm}
\paragraph{Answer sampling (Ans. Sampl.)}
Each (image,question) pair in the VQA dataset is annotated with 10 ground truth answers, corresponding to the different annotators. In those 10, we keep only the answers occuring more than 3 times, and randomly choose the one we ask our model to predict. 
\vspace{-0.4cm}
\paragraph{Data augmentation (DAVG)}
We use Visual Genome \cite{krishnavisualgenome} as a data augmentation to train our model, keeping only the example whose answer is in our vocabulary. This triples the size of our training set.
\vspace{-0.4cm}
\paragraph{Ensembling}
MUTAN (5) consist in an ensemble of five models trained on \emph{train+val} splits. We use 3 attentional MUTAN architectures with one trained with additional Visual Genome data. The 2 other models are instances of MLB, which can be seen as a special case of MUTAN. Details about the ensembling will be provided in the supplementary material.

\vspace{-0.4cm}
\paragraph{Results}

\begin{table}
\begin{tabularx}{\columnwidth}{l*{4}{X}X}
\toprule
& \multicolumn{4}{c}{\textit{test-dev}} & \textit{test-std} \\
& Y/N & No. & Other & All & All\\
\hline
SMem 2-hop \cite{XuS16} & 80.87 & 37.32 & 43.12 & 57.99 & 58.24\\
Ask Your Neur. \cite{malinowski16ijcv} & 78.39 & 36.45 & 46.28 & 58.39 & 58.43\\
SAN \cite{YangHGDS16} & 79.3 & 36.6 & 46.1 & 58.7 & 58.9 \\
D-NMN \cite{AndreasRDK16} & 81.1 & 38.6 & 45.5 & 59.4 & 59.4 \\
ACK \cite{CVPR16AMA} & 81.01 & 38.42 & 45.23 & 59.17 & 59.44 \\
MRN \cite{kim2016b} & 82.28 & 38.82 & 49.25 & 61.68 & 61.84\\
HieCoAtt \cite{LuYBP16} & 79.7 & 38.7 & 51.7 & 61.8 & 62.1 \\
MCB (7) \cite{fukui16mcb} & 83.4 & 39.8 & 58.5 & 66.7 & 66.5 \\
MLB (7) \cite{Kim2017} & 84.57 & 39.21 & 57.81 & 66.77 & 66.89 \\
MUTAN (3) & 84.54 & 39.32 & 57.36 & 67.03 & 66.96 \\
MUTAN (5) & \textbf{85.14} & \textbf{39.81} & \textbf{58.52} & \textbf{67.42} & \textbf{67.36} \\
\bottomrule
\end{tabularx}
\caption{\label{sota} MUTAN performance comparison  on the test-dev and test-standard splits VQA dataset; ($n$) for an ensemble of $n$ models.}
\end{table}

State-of-the-art comparison results are gathered in Table~\ref{sota}.
Firstly, we can notice that bilinear models, \textit{i.e.} MCB~\cite{fukui16mcb} and MLB~\cite{Kim2017} have a strong edge over other methods with a less powerful fusion scheme.

MUTAN outperforms all the previous methods with a large margin on \textit{test-dev} and \textit{test-std}.
This validates the relevance of the proposed fusion scheme, which models precise interactions between modalities. 
The good performances of MUTAN (5) also confirms its complementarity with MLB, already seen in section~\ref{sec:compafusion} without attention mecanism: 
MLB learns informative mono-modal projections, wheras MUTAN is explicitly devoted to accurately models bilinear interactions. 
Finally, we can notice that the performance improvement of MUTAN in this enhanced setup is conform to the performance gap reported in section~\ref{sec:compafusion}, 
showing that the benefit of the fusion scheme directly translates for the whole VQA task. 

Finally, we also evaluated an ensemble of 3 models based on the MUTAN fusion scheme (without MLB), that we denote as MUTAN (3). This ensemble also outperforms state-of-the-art results. We can point out that this improvement is reached with is an ensembling of 3 models, 
which is smaller than the previous state-of-the-art MLB results containing an ensembling of 7 models.

\subsection{Further analysis}

\begin{figure}
\includegraphics[width=\columnwidth]{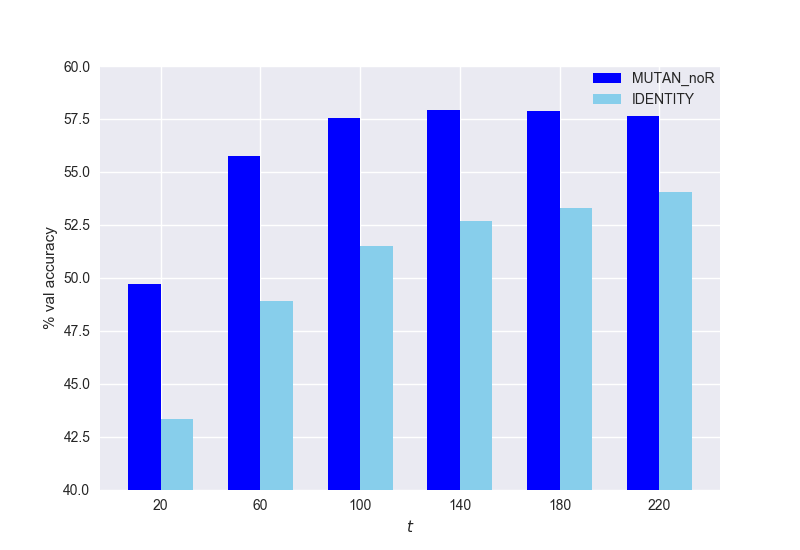}
\caption{\label{tucker_vs_mlb} The improvements given by MUTAN\_noR over a model trained with the identity tensor as a fusion operator between $\tq$ and $\tv$.}
\end{figure}

\begin{figure}
 \includegraphics[width=\columnwidth]{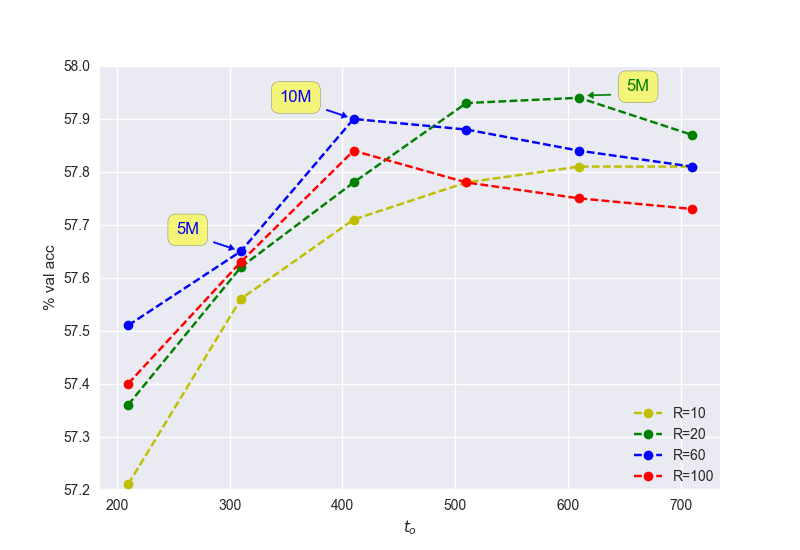}
\caption{\label{plot:r} Accuracy on VQA \textit{val} in function of $t_o$. Each colored dot shows the score of a MUTAN model trained on \textit{train}. The yellow labels indicate the number of parameters in the fusion.}
\end{figure}

\paragraph{Experimental setup}
In this section, we study the behavior of MUTAN under different conditions. Here, we examine under different aspects the fusion between $\q$ and $\vv$ with the Tucker decomposition of tensor $\tens{T}$. As we did previously, we don't use the attention mechanism in this section. We only consider a global visual vector, computed as the average of the $14 \times 14$ region vectors given by our CNN. We also don't use the answer sampling, asking our model to always predict the most frequent answer of the 10 ground truth responses. All the models are trained on the VQA \textit{train} split, and the scores are reported on \textit{val}.
\vspace{-0.3cm}
\paragraph{Impact of a plain tensor}
The goal is to see how important are all the parameters in the core tensor $\tens{T}_c$, which  model the correlations between projections of $\q$ and $\vv$. We train multiple MUTAN\_noR, where we fix all projection dimensions to be equal $t_q = t_v = t_o = t$ and $t$ ranges from 20 to 220. In Figure~\ref{tucker_vs_mlb}, we compare these MUTAN\_noR with a model trained with the same projection dimension, but where $\tens{T}_c$ is replaced by the 
identity tensor\footnote{This is strictly equivalent to MLB \cite{Kim2017} without attention. However, we are fully aware that it takes between 1000 and 2000 dimensions of projection to be around the operating point of MLB. With our experimental setup, we just focus on the effect of adding parameters to our fusion scheme.}. One can see that MUTAN\_noR gives much better results than identity tensor, even for very small core tensor dimensions. This shows that MUTAN\_noR is able to learn powerful correlations between modalities\footnote{Notice that for each $t$, MUTAN\_noR has $t^3$ parameters. For instance, for $t=220$, MUTAN adds 10.6M parameters over identity.}.
\vspace{-0.3cm}
\paragraph{Impact of rank sparsity}
We want to study the impact of introducing the rank constraint in the core tensor $\tens{T}_c$. We fix the input dimensions $t_q = 210$ and $t_v = 210$, and vary the output dimension $t_o$ for multiple rank constraints $R$. As we can see in Figure~\ref{plot:r}, controlling the rank of slices in $\tens{T}_c$ allows to better model the interactions between the unimodal spaces. The different colored lines show the behavior of MUTAN for different values of R. Comparing $R=60$ (blue line) and $R=20$ (green line), we see that a lower rank allows to reach higher values of $t_o$ without overfitting. The number of parameters in the fusion is lower, and the accuracy on the \textit{val} split is higher. 
\vspace{-0.3cm}
\paragraph{Qualitative observations}

\begin{figure}
\centering
\begin{subfigure}{0.49\linewidth}
\includegraphics[width=\linewidth]{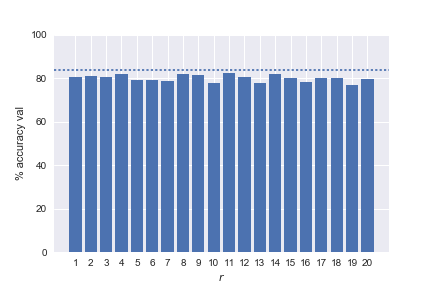}
\caption{"Is there"}
\end{subfigure}
%\hspace{1px}
\begin{subfigure}{0.49\linewidth}
\includegraphics[width=\linewidth]{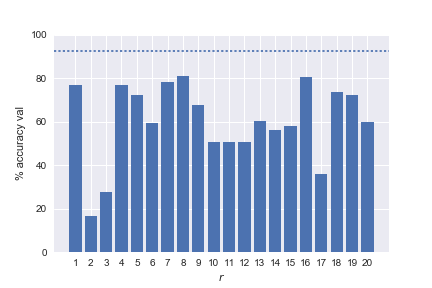}
\caption{"What room is"}
\end{subfigure}

\begin{subfigure}{0.49\linewidth}
\includegraphics[width=\linewidth]{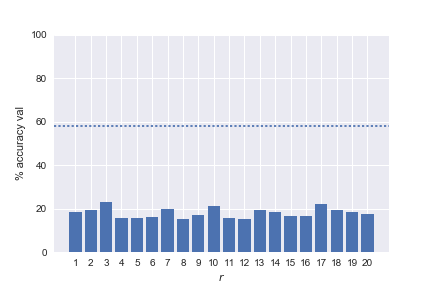}
\caption{"What is the man"}
\end{subfigure}
%\hspace{1px}
\begin{subfigure}{0.49\linewidth}
\includegraphics[width=\linewidth]{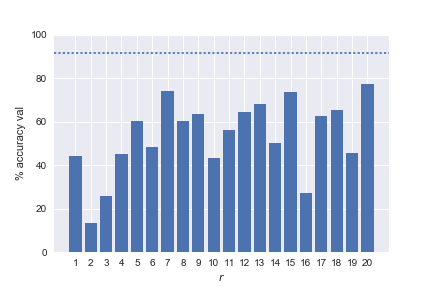}
\caption{"What sport is"}
\end{subfigure}
\caption{\label{rank_barplot} Visualizing the performances of ablated systems according to the R variables. Full system performance is denoted in dotted line.}
\end{figure}

\begin{figure}[]
\begin{subfigure}{\columnwidth}
\includegraphics[width=\linewidth]{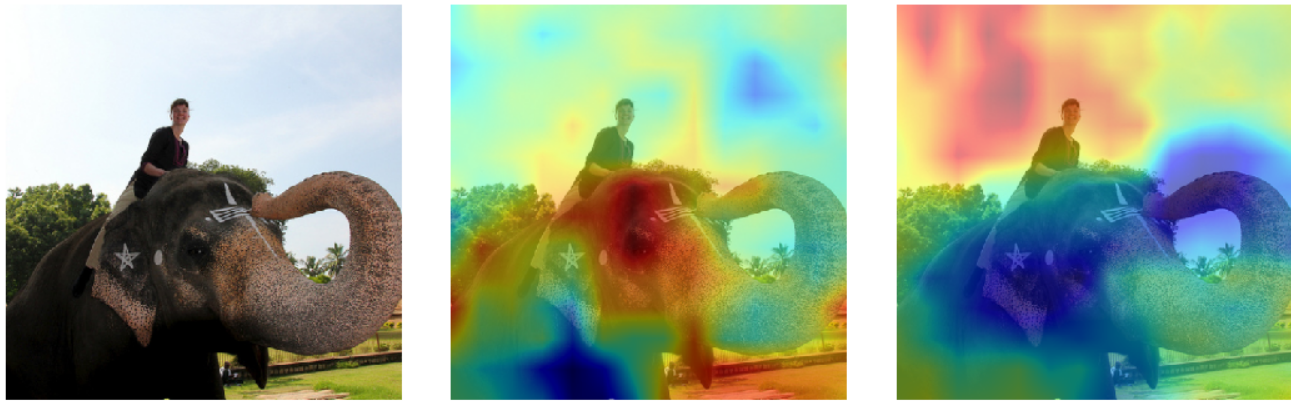}
\caption{Question: Where is the woman ? - Answer: on the elephant}
\end{subfigure}
\begin{subfigure}{\columnwidth}
\includegraphics[width=\linewidth]{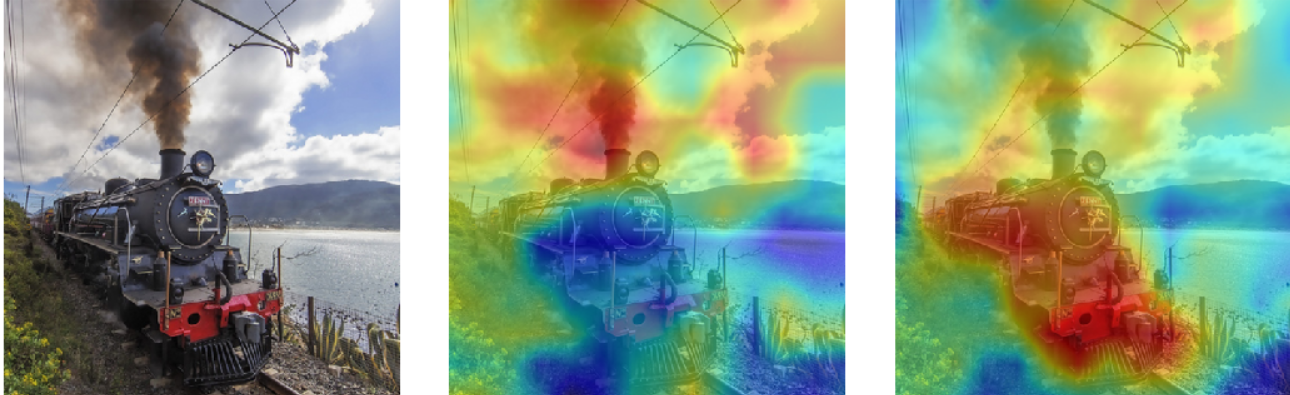}
\caption{Question: Where is the smoke coming from ? - Answer: train}
\end{subfigure}
\caption{\label{r_attention}The original image is shown on the left. The center and right images show heatmaps obtained when turning off all the projections but one, for two different projections. Each projection focuses on a specific concept needed to answer the question.}
\end{figure}

In MUTAN, the vector $\z$ that encodes the (image,question) pair is expressed as a sum over R vectors $\z_r$. We want to study the $R$ different latent projections that have been learnt during training, and assess whether the representations have captured different semantic properties of inputs. We quantify the differences between each of the R spaces using the VQA question types. We first train a model on the \textit{train} split, with $R=20$, and measure its performance on the \textit{val} set. Then, we set to 0 all of the $\z_r$ vectors except one, and evaluate this ablated system on the validation set. In Figure~\ref{rank_barplot}, we compare the full system to the R ablated systems for 4 different question types. The dotted line shows the accuracy of the full system, while the different bars show the accuracy of the ablated system for each R. Depending on the question type, we observe 3 different behaviors of the ranks. When the question type's answer support is small, we observe that each rank has learnt enough to reach almost the same accuracy as the global system. This is the case for questions starting by "Is there", whose answer is almost always "yes" or "no". Other question types require information from all the latent projections, as in the case of "What is the man". This leads to cases where all projections perform equally and significantly worst when taken individually than when combined to get the full model. At last, we observe that specific projections contribute more than others depending on the question type. For example, latent variable 16 performs well on "what room is", and is less informative to answer questions starting by "what sport is". The opposite behavior is observed for latent variable 17.

We run the same kind of analysis for the MUTAN fusion in the attention mechanism. In Figure~\ref{r_attention}, we show for two images the different attentions that we obtain when turning off all the projections but one. For the first image, we can see that a projection focuses on the elephant, while another focuses on the woman. Both these visual informations are necessary to answer the question "Where is the woman ?". The same behavior is observed for the second image, where a projection focuses on the smoke while another gives high attention to the train.

\section{Conclusion}
In this paper, we introduced our MUTAN strategy for the VQA task. Our main contribution is a multimodal fusion between visual and textual information using a bilinear framework. Our model combines a Tucker decomposition with a low-rank matrix constraint. It is designed to control the full bilinear interaction's complexity. 
MUTAN factorizes the interaction tensor into interpretable elements, and allows an easy control of the model's expressiveness. We also show how the Tucker decomposition framework generalizes the most competitive VQA architectures. MUTAN is evaluated on the most recent VQA dataset, reaching state-of-the-art.

{\small
\bibliographystyle{ieee}
\bibliography{egbib}
}

\title{Supplementary material}
\author{}

\maketitle

\section*{Preprocessing details}
\paragraph{Image}
As in \cite{fukui16mcb} (MCB) or \cite{Kim2017} (MLB), we preprocess the images before training our VQA models as follow. We load and rescal the image to 448. It is important to notice that we keep the proportion. Thus, 448 will be the size of the smaller edge. Then, we crop the image at the center to have a region of size $448 \times 448$. We normalize the image using the ImageNet normalization. Finally, we feed the image to a pretrained ResNet-152 and extract the features before the last Rectified Linear Unit (ReLU).

\paragraph{Question}

We use almost the same preprocessing as \cite{fukui16mcb} or \cite{Kim2017} for the questions.
We keep the questions which are associated to the 2000 most occuring answers.
We convert the questions characters to lower case and remove all the ponctuations. We use the space character to split the question into a sequence of words. Then, we replace all the words which are not in the vocabulary of our pretrained Skip-thoughts model by a special "unknown" word ("UNK").
Finally, we pad all the sequences of words with zero-padding to match the maximum sequence length of 26 words. We use TrimZero as in \cite{Kim2017} to avoid the zero values from the padding.

\section*{Optimization details}

\paragraph{Algorithm}
It is important to notice that we use the classical implementation of Adam \footnote{\url{https://github.com/torch/optim/blob/master/adam.lua}} with a learning rate of $10^{-4}$ unlike in \cite{fukui16mcb} or \cite{Kim2017}. In fact, we tried RMSPROP, SGD Nesterov and Adam with or without learning rate decay. We found that Adam without learning rate decay was more convenient and lead to the same accuracy.

\paragraph{Batch size}
During the optimization process, we use a batch size of 512 for the models without an attention modeling. For the others, we use a batch size of 100, because the models are more memory consuming.

\paragraph{Early stopping}
As in \cite{fukui16mcb} and \cite{Kim2017}, we use early stopping as a regularizer. During our training process, we save the model parameters after each epoch. To evaluate our model on the evaluation server, we chose the best epoch according to the Open Ended validation accuracy computed on the \textit{val} split when available.

As in \cite{fukui16mcb} and \cite{Kim2017}, for the models trained on the \textit{trainval} split, we use the \textit{test-dev} split as a validation set and are obliged to submit several times on the evaluation server. Note that we are limited to 10 submissions per day. In practice, we submit 3 to 4 times per models for epochs associated to training accuracies between 63\% to 70\%.

\section*{Ensemble details}

In table \ref{fusion_compare_sup:testdev}, we report several single models which compose our two ensembles. MUTAN(3) is made of a MUTAN trained on the \textit{trainval} split with 2 glimpses, an other MUTAN with 3 glimpses and a third MUTAN with 2 glimpses trained on the \textit{trainval} split with the visual genome data augmentation. All three have been trained with the same hyper-parameters besides the number of glimpses.

MUTAN(5) is made of the three same MUTAN models of MUTAN(3) and two MLB models which can be viewed as a special case of our Multimodal Tucker Fusion. The first MLB has 2 glimpses and was trained on the \textit{trainval} split.  It has been made available by the authors of \cite{Kim2017} \footnote{\url{https://github.com/jnhwkim/MulLowBiVQA/tree/master/model}}. The second MLB has 4 glimpses and was trained by ourself on the \textit{trainval} split with the visual genome data augmentation.

The final results of both ensembles are obtained by averaging the features extracted before the final Softmax layer of all their models.

\section*{Scores details}

In table \ref{fusion_compare_sup:testdev}, we provide the scores for each answer type processed on the \textit{val} and \textit{test-dev} splits.
In table \ref{sota_sup}, we provide the same scores for \textit{test-dev} and \textit{test-standard}.

\begin{table*}[t]
\begin{tabularx}{18cm}{l C*{4}{C} C*{4}{C}}
	\toprule
&&\multicolumn{4}{c}{\textit{test-dev}} & \multicolumn{4}{c}{\textit{val}} \\
\cmidrule(l){3-6} \cmidrule(l){7-10}
Model & $\Theta$ & Y/N & No. & Other & All & Y/N & No. & Other & All\\ 
\hline
Concat & 8.9 & 79.25 & 36.18 & 46.69 & 58.91 & 80.01 & 33.72 & 45.46 & 56.92\\ 
MCB & 32 & 80.81 & 35.91 & 46.43 & 59.40 & 81.61 & 33.94 & 45.14 & 57.39 \\ 
 MLB & 7.7 & \textbf{82.02} & 36.61 & 46.65 & 60.08 & \textbf{82.36} & 34.35 & 45.54 & 57.91\\ 
 MUTAN\_noR & 4.9 & 81.44 & 36.42 & 46.86 & 59.92 & 82.28 & 35.07 & 45.48 & 57.94\\ 
MUTAN & 4.9 & 81.45 & \textbf{37.32} & \textbf{47.17} & \textbf{60.17} & 82.07 & \textbf{35.16} & \textbf{46.03} & \textbf{58.16}\\
\hline
\hline
\textcolor{black}{MUTAN+MLB} & 17.5 & 82.29 & 37.27 & 48.23 & 61.02 & 82.59 & 35.21 & 46.84 & 58.76  \\

\bottomrule
\end{tabularx}
\caption{\label{fusion_compare_sup:testdev} Comparison between different fusion under the same setup on the \textit{test-dev} split. $\Theta$ indicates the number of learnable parameters (in million).}  
\end{table*}

\begin{table*}[t]
\begin{tabularx}{18cm}{l C*{4}{C} C*{4}{C}}
\toprule
& \multicolumn{4}{c}{\textit{test-dev}} & \multicolumn{4}{c}{\textit{test-std}} \\
\cmidrule(l){2-5} \cmidrule(l){6-9}
& Y/N & No. & Other & All & Y/N & No. & Other & All\\
\hline
SMem 2-hop \cite{XuS16} & 80.87 & 37.32 & 43.12 & 57.99 & 80.0 & 37.53 & 43.48 & 58.24\\
Ask Your Neur. \cite{malinowski16ijcv} & 78.39 & 36.45 & 46.28 & 58.39 & 78.24 & 36.27 & 46.32 & 58.43\\
SAN \cite{YangHGDS16} & 79.3 & 36.6 & 46.1 & 58.7 & - & - & - & 58.9 \\
D-NMN \cite{AndreasRDK16} & 81.1 & 38.6 & 45.5 & 59.4 & - & - & - & 59.4 \\
ACK \cite{CVPR16AMA} & 81.01 & 38.42 & 45.23 & 59.17 & 81.07 & 37.12 & 45.83 & 59.44 \\
MRN \cite{kim2016b} & 82.28 & 38.82 & 49.25 & 61.68 & 82.39 & 38.23 & 49.41 & 61.84 \\
HieCoAtt \cite{LuYBP16} & 79.7 & 38.7 & 51.7 & 61.8 & - & - & - & 62.1 \\
MCB (7) \cite{fukui16mcb} & 83.4 & 39.8 & 58.5 & 66.7 & 83.2 & 39.5 & 58.0 & 66.5 \\
MLB (7) \cite{Kim2017} & 84.54 & 39.21 & 57.81 & 66.77 & 84.61 & 39.07 & 57.79 & 66.89 \\
MUTAN (3) & 84.57 & 39.32 & 57.36 & 67.03 & 84.39 & 38.70 & 58.20 & 66.96 \\
MUTAN (5) & \textbf{85.14} & \textbf{39.81} & \textbf{58.52} & \textbf{67.42} & \textbf{84.91} & \textbf{39.79} & \textbf{58.35} & \textbf{67.36} \\
\bottomrule
\end{tabularx}
\caption{\label{sota_sup} MUTAN performance comparison  on the \textit{test-dev} and \textit{test-standard} splits VQA dataset; ($n$) for an ensemble of $n$ models.}
\end{table*}

\end{document}